\documentclass{article} 

\usepackage{iclr2026_conference,times}

\usepackage{amsmath,amsfonts,bm}

\def\eqref#1{equation~\ref{#1}}

\def\1{\bm{1}}

\DeclareMathAlphabet{\mathsfit}{\encodingdefault}{\sfdefault}{m}{sl}
\SetMathAlphabet{\mathsfit}{bold}{\encodingdefault}{\sfdefault}{bx}{n}

\usepackage{comment}
\usepackage{hyperref}
\usepackage{fontawesome5}
\usepackage[utf8]{inputenc}

\usepackage{enumitem}
\setlist[enumerate,itemize]{leftmargin=*, labelindent=0pt}
    \usepackage{siunitx}
\usepackage{url}
\usepackage{graphicx}
\usepackage{wrapfig}
\usepackage{booktabs}
\usepackage{enumitem}
\usepackage{tabularx} 
\usepackage{url}      
\usepackage[table]{xcolor}  

\usepackage[utf8]{inputenc}  
\usepackage[T1]{fontenc}     
\usepackage{lmodern}         
\usepackage[russian,english]{babel} 
\usepackage{CJKutf8}         

\definecolor{lightgreen}{RGB}{0,150,0}  
\definecolor{lightred}{RGB}{204,0,0}    

\definecolor{red10}{RGB}{252,174,145}   
\definecolor{red20}{RGB}{251,106,74}    

\newcommand{\towervision}{\textsc{TowerVision}}
\newcommand{\towervideo}{\textsc{TowerVideo}}

\newcommand{\towerp}{\textsc{Tower+}}
\newcommand{\visionblocks}{\textsc{VisionBlocks}}

\definecolor{tagorange}{HTML}{D35400} 
\definecolor{tagpurple}{HTML}{6C3483} 
\definecolor{taggreen}{HTML}{27AE60}  
\definecolor{tagblue}{HTML}{3498DB}

\renewcommand{\headrulewidth}{0.4pt}

\newcommand{\synthetictag}{\textcolor{tagorange}{Synthetic (Generated)}}
\newcommand{\translatedtag}{\textcolor{tagpurple}{Translated (Augmented)}}
\newcommand{\humantag}{\textcolor{taggreen}{Public Data}}

\title{
\rule{\textwidth}{1pt}
\centering
\raggedright
\towervision{}: Understanding and Improving Multilinguality in Vision-Language Models
}

\author{\small\textbf{André G. Viveiros\thanks{Joint first authors. Corresponding author: \texttt{andre.viveiros@tecnico.ulisboa.pt}.} $^{1,2}$ Patrick Fernandes$^{*1,2,3}$ Saul Santos$^{1,2}$} \\
\small\textbf{Sonal Sannigrahi$^{1,2}$ Emmanouil Zaranis$^{1,2}$ Nuno M. Guerreiro\thanks{Work done while at Unbabel.} $^{4}$ Amin Farajian$^{\dagger 5}$} \\
\small\textbf{Pierre Colombo$^{6}$ Graham Neubig$^{3}$ André F. T. Martins$^{\dagger 1,2,5,7}$} \\[0.6em] 
$^{1}$Instituto Superior Técnico, Universidade de Lisboa \quad
$^{2}$Instituto de Telecomunicações \quad \\
$^{3}$Carnegie Mellon University \quad
$^{4}$Sword Health \quad
$^{5}$TransPerfect \quad \\
$^{6}$MICS, CentraleSupélec, Université Paris-Saclay \quad 
$^{7}$ELLIS Unit Lisbon \quad \\
}

\iclrfinalcopy 
\begin{document}

\maketitle

\begin{flushleft}
\vspace{-1.4em}
\hspace{4em} 
\includegraphics[height=1em]{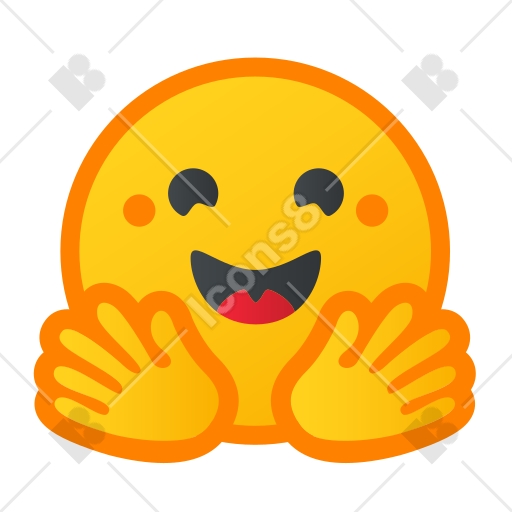}\ 
\href{https://huggingface.co/collections/utter-project/towervision-689a10be35396972889cadba}
{https://huggingface.co/collections/utter-project/towervision}
\\[0.01em] 
\hspace{4em} 
\includegraphics[height=1.19em]{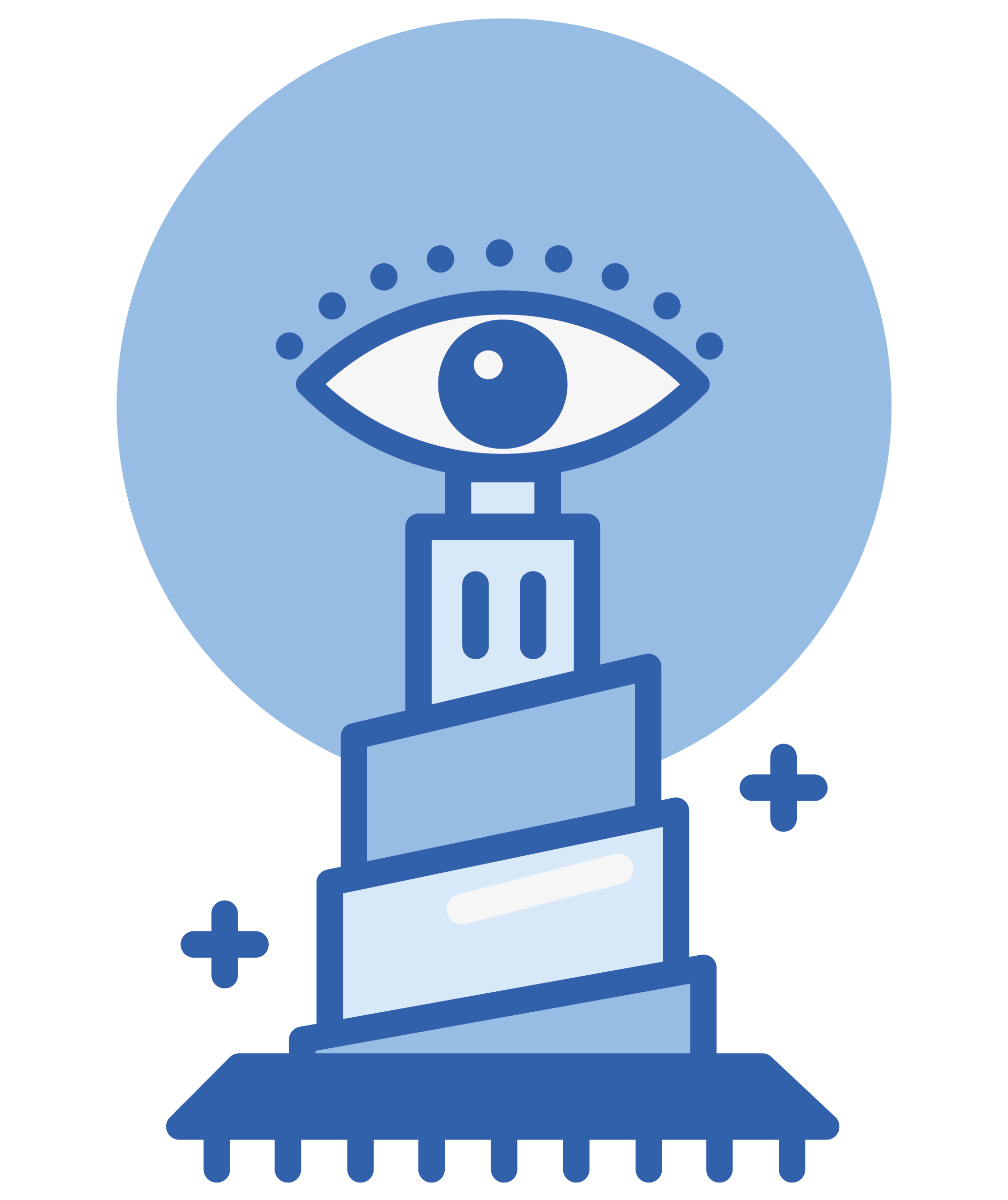}\ 
\href{https://guilhermeviveiros.github.io/TowerVision.io/}
{https://guilhermeviveiros.github.io/TowerVision.io/}
\end{flushleft}

\begin{abstract}

Despite significant advances in vision-language models (VLMs), most existing work follows an English-centric design process, limiting their effectiveness in multilingual settings. In this work, we 
provide a comprehensive empirical study analyzing the impact of several multilingual design choices, such as training data composition, encoder selection, and text backbones. The result is  
\towervision{}, a family of open multilingual VLMs for both image-text and video-text tasks, built upon the multilingual text-only model \towerp{}. \towervision{} achieves competitive performance on multiple multimodal multilingual benchmarks and shows particular strength in culturally grounded tasks and multimodal translation. By incorporating visual and cultural context during fine-tuning, our models surpass existing approaches trained on substantially larger datasets, as demonstrated on ALM-Bench and Multi30K (image tasks) and ViMUL-Bench (video tasks). 
Alongside the models, we release \visionblocks{}, a high-quality, curated vision-language dataset. 
Our findings highlight that multilingual vision-language training data substantially improves cross-lingual generalization---both from high-resource to underrepresented languages and vice versa---and that instruction-tuned LLMs are not always the optimal initialization point. 
To support further research, we publicly release all models, data, and training recipes.

\end{abstract}

\begin{center}
\vspace{-0.5cm} 
\includegraphics[width=0.7\textwidth]{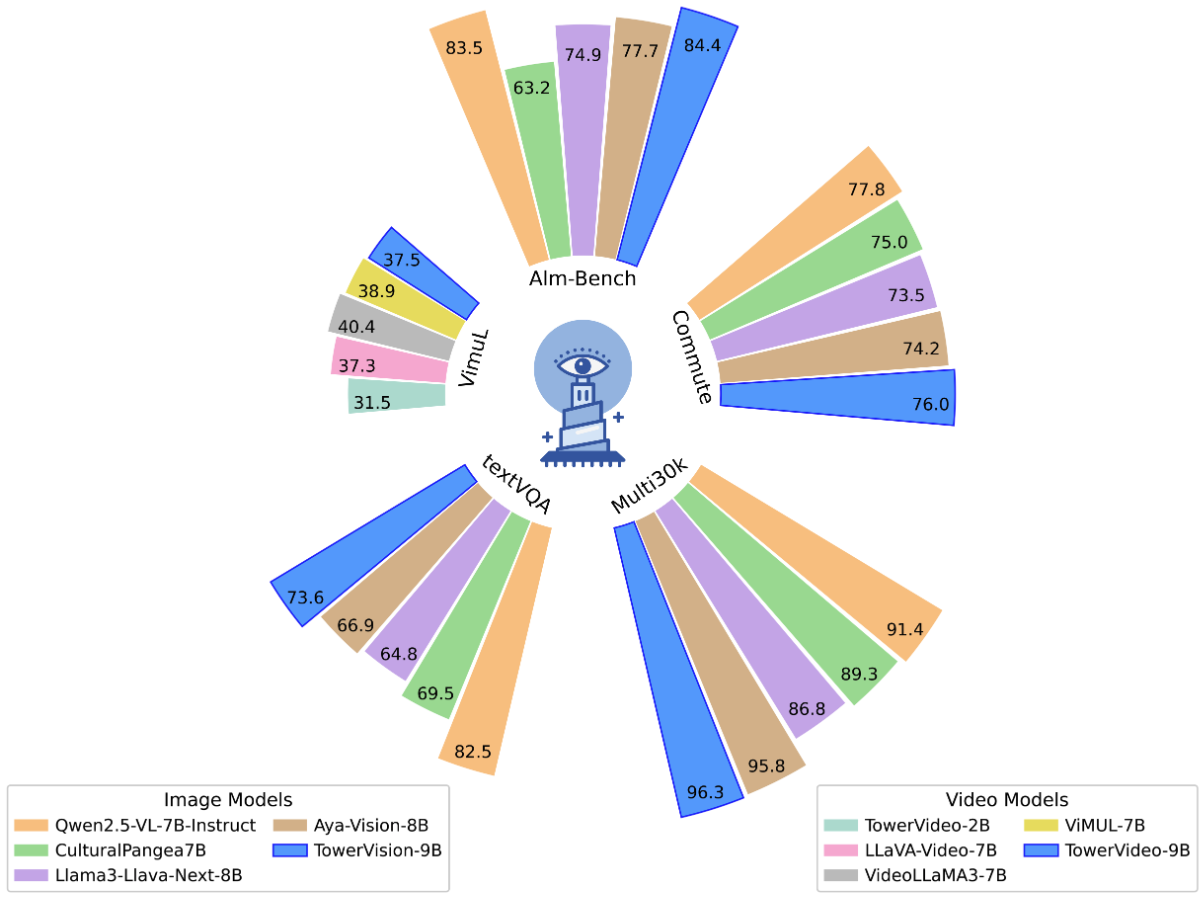}
\end{center}

\newpage

\section{Introduction}

The success and widespread adoption of large language models (LLMs) has naturally led to a surge of interest in adding multimodal capabilities to these models.
In particular, the visual modality has recently received considerable attention, with recent releases of \textit{frontier} vision-language models (VLMs) \citep{deitke2024molmopixmoopenweights,openai2024gpt4ocard,comanici2025gemini,team2025gemma,bai2025qwen25vl}. 
However, despite impressive progress, the development of VLMs has been mostly built upon English-centric language models, and trained with English vision-text data, giving little consideration to performance in most other languages. 
A key challenge in multilingualization of VLMs stems from an asymmetric data landscape---while high-quality \textit{text-only} multilingual corpora are relatively abundant, high-quality multilingual \textit{vision-text} data is scarce. As such, a critical challenge remains: What are the best strategies to effectively extend these models to support multiple languages beyond English? 



An effective strategy for VLM multilingualization is to leverage large-scale text-only multilingual  data to strengthen the language understanding of the text backbone, while complementing it with multimodal multilingual examples typically obtained through translation or high-quality synthetic generation, thereby reducing reliance on scarce real-world multilingual multimodal data. A recent example of this approach is \textsc{Pangea} \citep{yue2025pangeafullyopenmultilingual}, which extends multilinguality exclusively during the VLM fine-tuning stage using a mixture of data that combined multilingual vision-text pairs generated through synthetic data creation and machine translation of English instructions. While this strategy proved effective, it leaves open key questions: At which stages and on which modules should multilingualization be applied? Which design decisions yield the greatest impact? And how can visual grounding further enhance cross-lingual generalization?


\begin{figure*}[t]
    \centering
    \includegraphics[width=\textwidth]{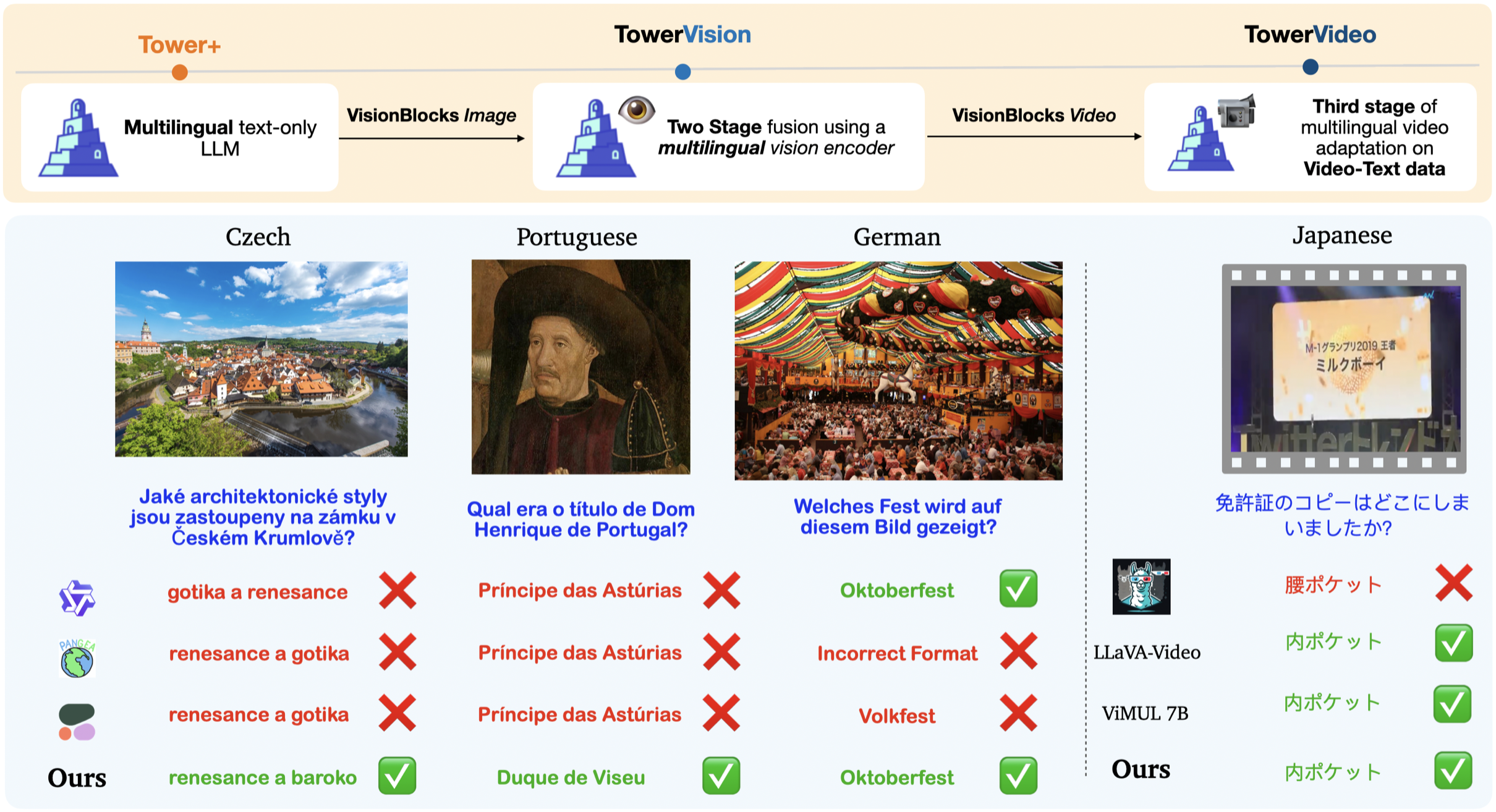}
    \caption{We present \towervision{} and \towervideo{}, open VLMs with enhanced cultural understanding and translation capabilities over leading open multimodal systems.}
    \label{fig:main-fig}
\end{figure*}

In this work, we introduce \towervision{}, \footnote{The TowerVision model collection is available \href{https://huggingface.co/collections/utter-project/towervision-689a10be35396972889cadba}{here}.} a suite of open-source multilingual VLMs built on top of \towerp{} models \citep{rei2025towerp} for 20 languages and dialects.%
\footnote{English, German, Dutch, Spanish (Latin America), French, Portuguese (Portugal), Portuguese (Brazilian), Ukrainian, Hindi, Chinese (Simplified), Chinese (Traditional), Russian, Czech, Korean, Japanese, Italian, Polish, Romanian, Norwegian (Nynorsk) and Norwegian (Bokmål)} %
To train \towervision{},  we systematically address the challenges outlined above through comprehensive ablation studies, component-level analysis, and cross-lingual evaluation of a multilingualization recipe. Specifically, we investigate how to enhance the multilingual capabilities of VLMs from two axes: first, by performing several ablations to understand and explore the impact of the underlying components (including the alignment projector, vision encoder and text-only LLM); and second, by creating better, more multilingual vision-text datasets and exploring the impact of using this data across different VLM training stages. Overall, compared to strong VLMs of similar size, \towervision{} exhibits competitive or superior performance on various multilingual and multimodal benchmarks, as well as cross-lingual transfer capabilities. 

In addition to image-based VLMs, we also train a separate multilingual video model,  \towervideo{}, 
built on top of \towervision{}, thereby extending our analysis to the video modality. \towervideo{} achieves competitive performance on  ViMUL-Bench \citep{shafique2025culturallydiversemultilingualmultimodalvideo}, a culturally-diverse multilingual video benchmark. 
Taken together, these contributions provide a comprehensive and systematic study of how to best integrate multilinguality into VLMs across modalities, architectural components, and training stages. Complementing the \towervision{} family, we also release \visionblocks{} \footnote{Released on Hugging Face upon acceptance.}, a curated dataset that consolidates and filters existing vision/video-language resources, further enriched with quality-controlled translations of English textual descriptions into 20 languages and dialects.


\section{\towervision{}}
\label{subsec:recipe}



Our approach follows a multi-stage process encompassing three key components, illustrated in Figure~\ref{fig:main-fig}: (i) a multilingual text-only backbone model, \towerp{} \cite{rei2025towerp}; (ii) a Vision Transformer encoder  (ViT; \citealt{dosovitskiy2021an}) that processes visual inputs and extracts meaningful features; (iii) a connector/adapter module that transforms these visual features to generate representations compatible with the text embedding space. 
These modules can be selectively trained or kept frozen during different stages of development~\citep{li2025surveystateartlarge}. 
Although this training recipe and variations thereof are well-established and have produced several high-quality models~(e.g., LLaVA \citep{liu2023visualinstructiontuning}, Intern-VL \citep{chen2024internvl}, NVLM \citep{dai2024nvlmopenfrontierclassmultimodal}, Qwen2.5-VL \citep{bai2025qwen25vl}, Molmo \citep{deitke2024molmopixmoopenweights}), most of these fall short in capturing multilingual and culturally diverse nuances. We therefore introduce our multilingual adaptation, \towervision{}---we first describe our carefully curated multilingual vision-text data, \visionblocks{} (\S\ref{sec:visionblocks}) and then describe the overall architecture and training procedure (\S\ref{sec:towervision}).

\subsection{\textsc{VisionBlocks}: Towards Better Multilingual Vision-Text Data}
\label{sec:visionblocks}

Creating a large-scale, high-quality, multilingual multimodal dataset for training visual language models to be helpful assistants is non-trivial for a series of intertwined reasons:

\begin{itemize}
    \item \textit{Human-written} vision-text data featuring user-model interactions (common in text-only alignment) is severely limited.
    While abundant data exists from large-scale captioning datasets (e.g., LAION-5B; \citealt{schuhmann2022laion5bopenlargescaledataset}), such sources over prioritize scale over quality which is not ideal for training VLMs with advanced capabilities \citep{dong2025scalablevisionlanguagemodel,zhou2023limaalignment} like instruction-following, helpfulness, and safety.
    \item High-quality  \textit{multilingual} vision-text data is scarce; furthermore, the lack of open, high-quality multilingual VLMs makes controlled synthetic data challenging or restricted to closed models with limited usage licenses. The most viable alternative, also employed by \textsc{Pangea}~\citep{yue2025pangeafullyopenmultilingual}, involves translating English vision-text interactions into target languages.
    \item Filtering techniques such as reward model scoring or LLM-as-judge approaches \citep{gu2025surveyllmasajudge} are significantly more challenging to implement for vision-text data, where even state-of-the-art VLMs (both open and proprietary) struggle to provide reliable preferences~\citep{li2024vlrewardbenchchallengingbenchmarkvisionlanguage}.
\end{itemize}

\begin{figure*}[t]
    \centering
    \includegraphics[width=0.9\textwidth]{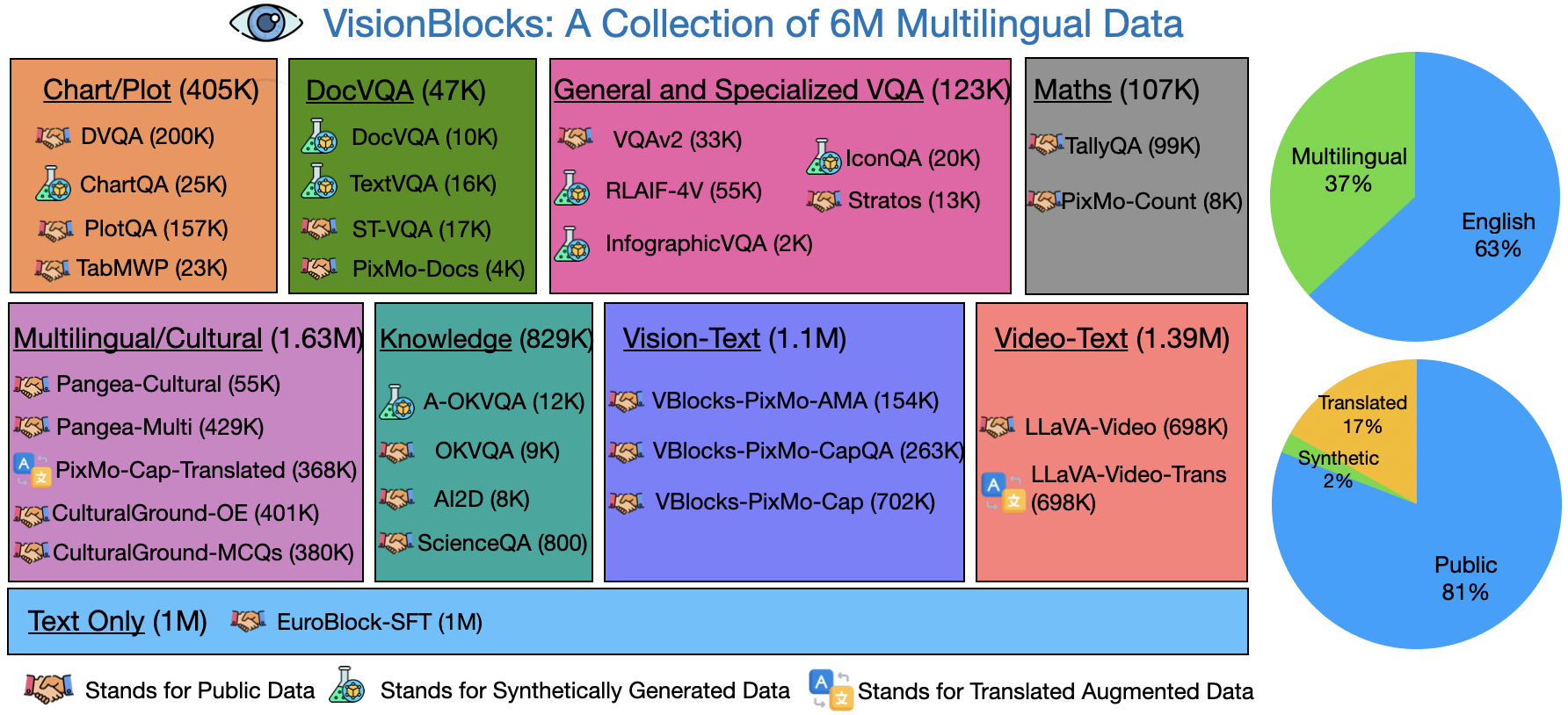}
    \caption{Overview of the \visionblocks{} dataset. See Table~\ref{tab:vision_blocks_full}, \S\ref{sec:app_visionblocks} for details.}
    \label{fig:vision_blocks}
\end{figure*}

With this in mind, we develop and release  \visionblocks{} (Figure~\ref{fig:vision_blocks}), which aggregates and filters data from multiple sources, enhanced with new translated and synthetic data, as described below. 

\paragraph{Collection of existing VLM data} 
For English vision-text data, we use the mixture created in \textsc{PixMo} ~\citep{deitke2024molmopixmoopenweights} with a few minor changes: we exclude the AndroidControl, Points, and PointQA datasets, as they do not provide additional multilingual value at this stage; 
For multilingual vision-text data, we leverage a subset of  ``Open-Ended'' and ``Multiple-Choice'' questions from \textsc{CulturalGround} \citep{nyandwi2025groundingmultilingualmultimodalllms} and the ``Cultural'' split of \textsc{PangeaIns} \citep{yue2025pangeafullyopenmultilingual} for our languages of interest. 
The samples from \textsc{PangeaIns} are originally found in  LAIONMulti~\citep{schuhmann2022laion5bopenlargescaledataset} that undergoes a series of automatic steps (using Gemini 1.5 Pro~\citep{geminiteam2024gemini15unlockingmultimodal}) including curating high-quality English instructions, carefully translating them to multiple languages, and adapting them for culturally-relevant multilingual contexts. 
\textsc{CulturalGround} uses a data curation pipeline that gathers culturally relevant entities from the Wikidata knowledge base, creates several questions and answers about each entity, rephrases them using an LLM, and filters low-quality samples using a VLM. 
In our work, we rely exclusively on \textsc{CulturalGround}'s filtered subsets to ensure maximum quality.

\paragraph{Translated and synthetic generated vision-language data} 

In addition to the original English and multilingual captions, 
we translate the highly curated \textsc{PixMo-Cap} caption data \cite{deitke2024molmopixmoopenweights} to our target languages using a \textsc{Tower} model~\citep{alves2024tower}. 
These translations are scored using \textsc{CometKiwi}  ~\citep{rei-etal-2022-cometkiwi} and filtered with a high threshold of 0.85 to ensure maximum quality. To further enhance diversity, we pair the remaining high-quality translations with a variety of language-specific captioning prompt templates (\S\ref{cap:tower_translation_captions}). 
We also augment the dataset with synthetic captions generated by the Gemini 2.5 API. For each image, we sample multiple system prompts to elicit diverse and detailed descriptions (see \S\ref{cap:gemini_synthetic_captions}). This augmentation is intended to improve coverage of fine-grained visual details (e.g., spatial relations, attributes, and contextual cues) that human-authored captions often omit, and provides instruction-like supervision, aligning our model more closely with recent VLM training paradigms that leverage synthetic data to boost generalization and response quality. Similar strategies have been shown to be effective in scaling up instruction-following capabilities of VLMs such as LLaVA~\citep{liu2023llava} and InstructBLIP~\citep{dai2023instructblipgeneralpurposevisionlanguagemodels}. 

\paragraph{Text-only data}
To retain text-only performance of the backbone LLM, while acquiring strong multimodal capabilities through multimodal training  requires a careful balance between text and multimodal data. Determining the optimal data mixture is non-trivial and typically involces a bunch of extensive ablation studies. For instance Pangea \citep{yue2025pangeafullyopenmultilingual} include approximately 10\% text-only data in their multimodal SFT mixture to retain text performance. In our work, we found that using around 20\% text-only data provided the best trade-off between text-only and multimodal benchmark performance. We used \textsc{EuroBlocks} as the extended text-only data, a curated multilingual collection of high-quality synthetic data from the \textsc{EuroLLM} \citep{martins2025eurollm}  synthetic post-training data. \textsc{EuroBlocks} provides diverse, instruction-aligned text that enriches our dataset with robust multilingual coverage and fine-grained, high-quality descriptions.

\paragraph{Translated multilingual video data}
As video-text data, we employ the LLaVA-Video-178k dataset \citep{zhang2025llavavideovideoinstructiontuning}, which contains captions alongside open-ended and multiple-choice questions in English. To make the dataset multilingual, we retain a randomly sampled half of the conversations in English, and we translate the remaining half uniformly into all supported languages using \towerp 9B \citep{rei2025towerp}, thereby ensuring balanced cross-lingual coverage.

\subsection{\towervision{}: Architecture \& Training Details}\label{sec:towervision}

One way to improve the multilinguality of LLMs (e.g., improving cross-lingual understanding or extending multilingual support for other languages) is to start from a strong pretrained model and continue pretraining on carefully curated data, with subsequent post-training \citep{xu2024paradigm,xu2025xalma,alves2024tower}. 
\towervision{} follows a similar principle,  starting from a strong multilingual Gemma-based backbone \towerp{} 2B/9B \citep{rei2025towerp}, which achieves strong multilingual general-purpose performance by leveraging a curated high-quality multilingual dataset and a training recipe designed to preserve general capabilities. As shown in \S\ref{tab:backbone_ablation}, starting from this multilingual backbone substantially improves cross-lingual performance compared to starting from Gemma.

For the vision encoder,  \towervision{} is initialized with the recently proposed SigLIP2-so400m/14@384px \citep{tschannen2025siglip2multilingualvisionlanguage}, a vision transformer operating at $384\times384$ resolution that extracts image patch representations and produces multilingually-aligned embeddings of size $729$. SigLIP2 is trained on a more diverse data mixture compared to alternatives such as  CLIP-ViT \citep{radford2021learningtransferablevisualmodels}, Perception Encoder \citep{bolya2025perceptionencoderbestvisual}, or SigLIP1 \citep{zhai2023sigmoidlosslanguageimage}, and thereby yields better multilingual understanding, as we shall see in \S\ref{sec:ablations}. To align the vision and text modalities, we use a LLaVA-based architecture \citep{liu2023visualinstructiontuning}, where we train a projection layer consisting of a 2-layer MLP, randomly initialized.
By combining \towerp{} for text and SigLIP2 for vision, \towervision{} benefits from complementary multilingual strengths across both modalities. The training process consists of three stages:

\begin{enumerate}
    \item A \textit{projector pretraining} phase, where we train the model to predict captions given images on the \textsc{PixMo}-Cap dataset, freezing both the vision encoder and the language model backbone (so only the projector is trained). Each image is encoded once (downscaled to 384$\times$384 if necessary). During this phase, we focus exclusively on diverse, high-quality English captions, which we show to be more effective for aligning visual and textual representations (see \S\ref{sec:ablations}). 
    \item A \textit{vision finetuning} phase, where we unfreeze the full model 
    and train it on the full \textsc{VisionBlocks} dataset (\S\ref{sec:visionblocks}), excluding the video-text data. In this phase, we use \textit{high-dynamic resolution} \citep{liu2024llavanext}, breaking high-resolution images into a grid of smaller tiles which are then encoded with the vision encoder independently (together with a global thumbnail tile). The projected embeddings are then concatenated. We use a maximum of six tiles, which provides the best trade-off 
    (\S\ref{cap:tiles_patch_vision_encoders}). 
    This phase leads to the \towervision{} model. 
    \item A \textit{video finetuning} phase, where the video portion of \visionblocks{} is used to finetune   \towervision{}  on 32-frame video inputs at the encoder’s fixed resolution of 384$\times$384. Unlike the previous stage, we omit tiling for efficiency. 
    This phase leads to the \towervideo{} model. 
\end{enumerate}

The models were trained on a custom fork of the LLaVA-Next~\citep{liu2024llavanext} codebase.%
\footnote{https://github.com/deep-spin/LLaVA-NeXT} 

\section{Evaluation \& Main Results}

We evaluate \towervision{} and \towervideo{} on a comprehensive suite of benchmarks spanning multiple modalities and task types (single-image, multi-images, and video) across diverse languages, both within and beyond our training set. In this section, we focus on vision-language tasks, which includes multilingual visual/video question answering, cultural understanding, OCR-related tasks, and visual-language understanding, as well as multilingual video-language tasks. Our assessment relies primarily on closed-form tasks, complemented by large language models serving as judges for video-based evaluations.

\subsection{Tasks \& Evaluation Benchmarks}

\paragraph{Vision-language tasks} 
We report results on ALM-Bench \citep{vayani2024alm},  
a cultural understanding multilingual\footnote{German, Spanish, French, Italian, Korean, Dutch, Russian, English, Portuguese, Chinese (Simplified and Traditional), Icelandic, Czech, Ukrainian, Hindi, Japanese, Polish, Swedish, Hungarian, Romanian, Danish, Norwegian (Nynorsk), and Finnish.} visual QA benchmark, OCRBench~\citep{Liu_2024} and cc-OCR~\citep{yang2024ccocrcomprehensivechallengingocr} for English and multilingual\footnote{German, French, Italian, Russian, Spanish, Korean, Portuguese.} OCR-centric capabilities respectively, and TextVQA~\citep{singh2019text_vqa}, assessing scientific understanding. Within cc-OCR, we report results on the multilingual text reading subset, as our primary focus is to evaluate the model's multilingual text recognition capabilities.

\paragraph{Multimodal translation} 
We report results on CoMMuTE~\citep{futeral-etal-2023-tackling}, a specialized multimodal translation benchmark that uses the visual content to resolve lexical ambiguities present in the source language, and Multi30K~\citep{elliott-etal-2016-multi30k}, a standard benchmark for multimodal machine translation (MT) of image captions. 

\paragraph{Culturally-aware multilingual video tasks} We use ViMUL-Bench  \citep{shafique2025culturallydiversemultilingualmultimodalvideo}, a multilingual video QA benchmark spanning 14 languages: Arabic (ar), Bengali (bn), Chinese (zh), English (en), French (fr), German (de), Hindi (hi), Japanese (ja), Russian (ru), Sinhala (si), Spanish (es), Swedish (sv), Tamil (ta), and Urdu (ur). The dataset contains both open-ended and multiple-choice questions covering culturally diverse domains such as festivals, customs, food, and heritage. Unlike prior datasets, ViMUL-Bench enables comprehensive evaluation of video-language models across both high- and low-resource languages, promoting inclusive and culturally aware research.


\subsection{Baselines}
For evaluation, we leverage the lmms-eval framework~\citep{lmms_eval2024}, which enables a systematic comparison of \towervision{} against leading open VLMs. We include several multilingual multimodal models, such as \textit{CulturalPangea-7B}~\citep{yue2025pangeafullyopenmultilingual}, designed to address gaps in multilingual cultural understanding, and \textit{Aya-Vision-8B} \citep{singh2024aya},  optimized for a broad range of vision-language tasks. 
In addition, we evaluate models from the \textit{Gemma3-Instruct}  (\textit{Gemma3-4B-it}, \textit{Gemma3-12B-it}; \citealt{team2025gemma}) and the \textit{Qwen2.5-VL-Instruct} families (\textit{Qwen2.5-VL-3B-Instruct}, \textit{Qwen2.5-VL-7B-Instruct}; \citealt{qwen2025qwen25technicalreport}), both of which have demonstrated strong performance across a variety of multimodal benchmarks. Finally, we report results for a LLaVA-based model, \textit{Llama3-Llava-Next-8B}~\citep{liu2024llavanext}, a general-purpose VLM with strong performance across a wide range of tasks. The exact checkpoints for all models are listed in \S\ref{app:model_checkpoints}.

 For \towervideo{}, we consider several competitive open-source video models of comparable scale, including \textit{VideoLLaMA3-7B} \citep{damonlpsg2025videollama3}, \textit{LLaVA-Video-7B} \citep{zhang2025llavavideovideoinstructiontuning} also trained on LLaVA-Video-178k, and \textit{ViMUL-7B} \citep{shafique2025culturallydiversemultilingualmultimodalvideo}, a multilingual video model.

\begin{table}[t]
\caption{\textbf{Vision-Language Model Performance.} English and multilingual VLMs results across multiple benchmarks. Reported values correspond to final accuracy ($\uparrow$).}
\vspace{0.2cm}
    \centering
    \small
    \resizebox{\textwidth}{!}{%
    \begin{tabular}{lccccc}
    \toprule
    & \multicolumn{2}{c}{\textbf{English} ($\uparrow$)} & \multicolumn{3}{c}{\textbf{Multilingual} ($\uparrow$)} \\
    \cmidrule(lr){2-3} \cmidrule(lr){4-6}
    & TextVQA & OCRBench & CC-OCR & 
    ALM-Bench (en) & ALM-Bench (multi)
    \\
    \midrule
    Qwen2.5-VL-3B-Instruct  & 77.8 & 78.7 & 76.4 & 81.0 & 76.2 \\
    Qwen2.5-VL-7B-Instruct   & \textbf{82.5} & \textbf{84.5} & \textbf{78.6} & 83.1 & 83.6 \\
    Gemma3-4B-it & 65.2 & 74.2 & 69.1 & 79.7 & 80.0 \\
    Gemma3-12B-it  & 73.2 & 74.7 & 73.8 & 83.5 & 84.5  \\
    CulturalPangea7B & 69.8 & 63.5 & 51.7 & 61.3 & 65.2 \\
    Llama3-Llava-Next-8B  & 64.8 & 54.4 & 40.9 & 76.5 & 73.4  \\
    Aya-Vision-8B  & 66.9 & 61.0 & 46.3 & 78.2 & 77.3 \\
    \midrule
    \rowcolor{blue!10} TowerVision-2B & 68.1 & 58.6 & 46.1 & 77.1 & 81.1 \\
    \rowcolor{blue!20} TowerVision-9B  & 73.6 & 69.7 & 56.3 & \textbf{83.6} & \textbf{85.2} \\
    \bottomrule
    \end{tabular}}
    \label{tab:vlm_tasks}
\end{table}

\begin{table}[t]
    \caption{\textbf{Multimodal Translation Benchmarks.} 
    We report {\sc xComet} \citep{guerreiro2024xcomet} for Multi30K  and contrastive pairwise accuracy for CoMMuTE.}
    \vspace{0.2cm}
    \centering
    \small
    \resizebox{\textwidth}{!}{%
    \begin{tabular}{lcccccccc}
    \toprule
    & \multicolumn{3}{c}{Multi30K ($\uparrow$)} & \multicolumn{4}{c}{CoMMuTE ($\uparrow$)} \\
    \cmidrule(lr){2-4} \cmidrule(lr){5-8}
    & en$\rightarrow$cs & en$\rightarrow$de & en$\rightarrow$fr &en$\rightarrow$de & en$\rightarrow$fr & en$\rightarrow$ru & en$\rightarrow$zh \\ 
    \midrule
    Qwen2.5-VL-3B-Instruct & 83.3 & 96.7 & 92.6 & 71.6 & 74.4 & 77.5 & 81.5 \\
    Qwen2.5-VL-7B-Instruct & 83.9 & 97.1 & 93.2 &  74.7 & 76.9 & 77.2 & \textbf{82.4} \\
    Gemma3-4B-it & 33.4 & 44.0 & 33.2 & \textbf{76.7} & 78.2 & \textbf{79.0} & 74.4 \\
    CulturalPangea7B & 80.0 & 95.8 & 92.1 & 68.3 & 77.3 & 75.3 & 79.3 \\
    Llama3-Llava-Next-8B  & 79.1 & 93.3 & 88.1 & 72.0 & 74.4 & 74.4 & 73.5\\
    Aya-Vision-8B  & 94.4 & 97.9 & 95.3 & 69.3 & 76.9 & 74.4 & 76.2\\
    \midrule
    \rowcolor{blue!10}\towervision-2B & 90.3 & 97.5 & 94.7 & 70.0 & 74.3 & 73.2 & 76.6   \\
    \rowcolor{blue!20}\towervision-9B & \textbf{95.1} & \textbf{98.1} & \textbf{95.6} & 72.0 & \textbf{78.8} & 75.6 & 77.4\\
    \bottomrule
    \end{tabular}
    }
    \label{tab:multimodal_mt}
\end{table}

\begin{table*}[t]
\centering
\small
\caption{\textbf{Multilingual video performance per language.} Accuracy (\%) on ViMUL-Bench across 14 languages averaged across multiple-choice and open-ended questions. \underline{Underlined} values mark the best score within \towervision{}/\towervideo{} variants; Unsupported languages are marked with $^*$.}
\vspace{0.2cm}
\setlength{\tabcolsep}{3pt} 
\resizebox{\textwidth}{!}{%
\begin{tabular}{lcccccccccccccc}
\toprule
\textbf{Model} & \textbf{ar} & \textbf{bn$^*$} & \textbf{zh} & \textbf{en} & \textbf{fr} & \textbf{de} & \textbf{hi} & \textbf{ja} & \textbf{ru} & \textbf{si$^*$} & \textbf{es} & \textbf{sv} & \textbf{ta$^*$} & \textbf{ur$^*$} \\
\midrule
ViMUL-7B & 41.5 & 35.4 & 37.0 & 48.6 & 48.3 & 43.9 & \textbf{39.2} & 37.8 & 45.7 & 21.2 & 44.3 & 41.4 & 23.3 & \textbf{36.8} \\
LLaVA-Video-7B & 38.8 & 30.4 & 43.2 & \textbf{53.3} & \textbf{49.2} & 45.4 & 34.2 & 33.4 & 38.2 & 18.1 & 45.7 & 39.8 & 21.9 & 33.8 \\
VideoLLaMA3-7B & \textbf{45.6} & \textbf{36.6} & \textbf{48.0} & 52.9 & 47.1 & 43.8 & 37.5 & 39.4 & 44.8 & \textbf{25.1} & 45.4 & 38.5 & 22.8 & 32.1  \\
\midrule
\rowcolor{blue!5} \towervision-2B & 18.9 & \underline{19.5} & 21.7 & 34.2 & 28.9 & 28.3 & 25.1 & 22.2 & 24.8 & 16.3 & 30.4 & 27.1 & 16.1 & \underline{19.9} \\
\rowcolor{blue!10} \towervideo-2B & \underline{23.0} & 18.9 & \underline{35.9} & \underline{45.2} & \underline{39.6} & \underline{39.7} & \underline{37.2} & \underline{34.1} & \underline{38.0} & \underline{17.1} & \underline{37.4} & \underline{38.0} & \underline{17.7} & 18.7 \\
\midrule
\rowcolor{blue!15} \towervision-9B & 34.2 & \underline{25.4} & 35.3 & 46.7 & 41.1 & 40.8 & \underline{34.2} & 28.1 & 40.3 & 19.8 & 40.5 & 39.6 & 21.6 & 26.4\\
\rowcolor{blue!20} \towervideo-9B & \underline{38.6} & 22.1 & \underline{44.8} & \underline{51.9} & \underline{49.1} & \underline{\textbf{47.1}} & 32.2 & \underline{\textbf{42.3}} & \underline{40.9} & \underline{20.8} & \underline{\textbf{46.0}} & \underline{\textbf{44.8}} & \underline{\textbf{24.1}} & 19.5\\
\bottomrule
\end{tabular}
}
\label{tab:video_multilingual_performance}
\end{table*}
\subsection{Main Results}

Tables \ref{tab:vlm_tasks}--\ref{tab:multimodal_mt} report the performance of \towervision{} on vision-language benchmarks as well as multimodal translation benchmarks, while Table \ref{tab:video_multilingual_performance} reports the results on the multilingual video-language benchmark. 
We summarize the main findings below. 

\textbf{\towervision{} models are strong in cultural-aware tasks.} Within our suite of vision-language benchmarks, we achieve state-of-the-art results on ALM-Bench (Table \ref{tab:vlm_tasks}, a culturally diverse benchmark, in both the English and multilingual split. Qwen2.5VL-7B and Gemma3-12B are the closest competitors, while other baselines lag behind. In the multilingual split, we evaluate on a diverse set of 23 languages covering several language families and scripts. \towervision{} is able to exhibit enhanced cultural multimodal understanding, suggesting that it is still performant in less seen and unseen languages within its training data. We further assess the cross-lingual generalization capabilities of \towervision{} in \S\ref{tab:xcross_gen}.

\textbf{\towervision{} is less competitive on OCR-related tasks.} 
We hypothesize this is likely due to the limited amount of OCR-focused data in \visionblocks{} compared against other models. Since we primarily pretrained \towervision{} on large-scale image-caption datasets emphasizing natural images and language alignment, it struggles with scanned text or OCR-heavy scenarios. Despite these limitations, \towervision{} does obtain superior performance compared against AyaVision-8B and Llam3LLaVANext-8B, the former of which has seen significant amounts of OCR-specific data \citep{singh2024aya}.

\textbf{\towervision{}-2B is competitive multilingually with larger models.} 
In multimodal translation benchmarks, \towervision{} consistently demonstrates strong performance on Multi30K and is competitive on CoMMuTE (Table~\ref{tab:multimodal_mt}). Our 9B variant achieves state-of-the-art results on Multi30k across all language pairs, and we observe that even our smaller 2B variant is a competitive model against the larger baselines on translation-specific, as well as vision-language benchmarks. For instance, on Multi30K, \towervision{}-2B obtains superior scores to Qwen2.5VL-7B and CulturalPangea-7B. Similarly, on the multilingual split of ALM-Bench, \towervision{} 2B is competitive with Qwen2.5VL-7B and outperforms AyaVision-8B. These results further highlight the efficacy of \towervision{}’s multilinguality and design choices. We also note that scaling from 2B to 9B parameters consistently improves performance across all benchmarks, suggesting that our training recipe scales well.

\textbf{Multilingual  fine-tuning improves cross-lingual performance in \towervideo{}.} In Table~\ref{tab:video_multilingual_performance}, we report averages across multiple-choice accuracy and open-ended responses, which are automatically judged using GPT-4o \citep{openai2024gpt4ocard}, with the same evaluation prompt as \citet{shafique2025culturallydiversemultilingualmultimodalvideo}. We compare our \towervideo{} models, including the 9B variant, to strong open-source baselines. Our multilingual models are competitive across several languages despite using smaller datasets and fewer frames (for instance, VideoLLaMA3 uses 180 frames). Specifically, ViMUL was trained with separate copies of the dataset for each language, whereas our approach uses a single copy with half in English and the other half uniformly translated into the supported languages. Overall, these results highlight the effectiveness of video-based multilingual fine-tuning in improving cross-lingual reasoning. 

Overall, our results demonstrate the effectiveness of our design choices in endowing our model with strong multilingual capabilities due to a combination of increased multilingual culturally-sensitive training data, a more multilingual text backbone (\towerp{}), and a multilingual vision encoder. We detail these choices in \S\ref{sec:ablations} with a carefully conducted set of ablation experiments.

\section{Where and How Does Multilinguality Matter?}\label{sec:ablations} 

Following the main results of \towervision{}, we delve deeper into its design choices. 



\textbf{Multilingual backbones improve cross-modal performance.} 
\label{backbone_ablation}  
The choice of backbone in \towervision{} can substantially influence performance across multilingual and multimodal tasks. 
We focus on two complementary aspects. First, we examine the significance of multilingual capacity by comparing the \towerp{} backbone, which is highly multilingual and designed for general-purpose multilingual text tasks, against \textsc{Gemma2}, the model on which \towerp{} was built. 
Second, we investigate the impact of instruction tuning before modality fusion, which is widely applied in modern VLMs from the start \citep{liu2023visualinstructiontuning,bai2025qwen25vltechnicalreport}, but whose 
effect on the final model remains unclear. 
To study these effects, we train \towervision{} at 2B and 9B scales using four backbones: \textsc{Gemma2}-pt (pretrained, not instruction-tuned), \textsc{Gemma2}-it (instruction-tuned), \towerp{pt} (pretrained \towerp{}), and \towerp{it} (instruction-tuned \towerp{}), following the recipe in \S\ref{subsec:recipe}. 
As shown in Table \ref{tab:backbone_ablation}, using \towerp{} consistently outperforms \textsc{Gemma2}, confirming the importance of a strong multilingual backbone for robust cross-modal understanding. At smaller scales, non-instructed models (\textsc{Gemma2}-pt, \towerp{pt}) retain stronger raw visual extraction, while instruction-tuned variants excel in cultural knowledge and reasoning. By the 9B scale, this gap narrows, with instruction-tuned models integrating both skills and achieving state-of-the-art performance. These findings underscore the complementary roles of multilingual pretraining and instruction tuning, and the need for careful backbone selection in VLMs.

\begin{table}[t]
   \caption{\textbf{Impact of backbone and instruction tuning.}
    Performance of VLMs with different backbones on English and multilingual tasks. 
    }
    \vspace{0.2cm}
    \centering
    \small
    \resizebox{\textwidth}{!}{
    \begin{tabular}{lccccc}
    \toprule
    \textbf{Backbone Model}
    & \multicolumn{2}{c}{\textbf{English} ($\uparrow$)} & \multicolumn{3}{c}{\textbf{Multilingual} ($\uparrow$)} \\
    \cmidrule(lr){2-3} \cmidrule(lr){4-6}
    & TextVQA & OCRBench & CC-OCR & 
    ALM-Bench (en) & ALM-Bench (multi) \\   
    \midrule
    {\sc Gemma2}-pt-2B & 69.2 & 61.2 & 45.3 & 74.3 & 76.7 \\
    \rowcolor{blue!10} 
    \towerp{pt}-2B  & \textbf{70.3} & 62.1 & \textbf{46.3} & 73.0 & 78.2 \\ 
    {\sc Gemma2}-it-2B & 70.0 & \textbf{63.0} & 45.9 & 75.0 & 75.1 \\ \rowcolor{blue!10} 
    \towerp{it}-2B  & 68.1 & 58.6 & 46.1 & \textbf{77.1} & \textbf{81.1} \\ 
    \midrule
    {\sc Gemma2}-pt-9B & 72.4 & 66.6 & 49.6 & 79.9 & 79.6 \\
    \rowcolor{blue!15} 
    \towerp{pt}-9B & 73.2 & 64.5 & 54.5 & 81.3 & 84.4 \\
    {\sc Gemma2}-it-9B & \textbf{74.4} & 67.2 & 49.5 & 79.6 & 81.5 \\   \rowcolor{blue!20} 
    \towerp{it}-9B & 73.6 & \textbf{69.7} & \textbf{56.3} & \textbf{83.6} & \textbf{85.2} \\
    \bottomrule
    \end{tabular}
    }
    \label{tab:backbone_ablation}
\end{table}

\textbf{Multilingual-aware vision encoders improve performance in low-data regimes.} 
Effectively leveraging multilingual data is crucial for VLMs, yet it is unclear whether the vision encoder’s own multilingual capacity plays an important role. 
We compare SigLIP2, trained on diverse multilingual data, with SigLIP1, an earlier English-centric version, to test whether multilingual-aware encoders are essential or if sufficient fine-tuning can compensate. 
We train \towervision{} with both encoders on English-only and multilingual data at 2B and 9B scales (results in Table~\ref{tab:siglip2_vs_siglip1}). 

\begin{wraptable}{r}{8.1cm}
\vspace{-0.7cm}
\centering
\small
    \centering
    \caption{Multilingual impact of different vision encoders.}
    \vspace{0.2cm}
    
    \begin{tabular}{lcccc}
    \toprule
     \textbf{\towervision{}} & \multicolumn{2}{c}{\textbf{2B}} & \multicolumn{2}{c}{\textbf{9B}} \\
    \cmidrule(lr){2-3} \cmidrule(lr){4-5}
    \textbf{Variant} 
     & \textbf{En} & \textbf{Multi} & \textbf{En} & \textbf{Multi} \\
    \midrule
    SigLIP1-En & 67.4 & 60.2 & 78.3 & 81.2 \\
    SigLIP2-En & 69.3 & 67.1 & 77.2 & 81.1 \\
    SigLIP1-(En+Multi) & 76.6 & 80.7 & 83.6 & 84.4 \\
    SigLIP2-(En+Multi) & \textbf{77.1} & \textbf{81.1} & \textbf{83.6} & \textbf{85.2} \\
    \bottomrule
    \end{tabular}
    \label{tab:siglip2_vs_siglip1}
\end{wraptable}

Without additional multilingual data, SigLIP2 models consistently outperform SigLIP1, showing clear benefits in low data regimes. With multilingual fine-tuning, however, the gap narrows, showing that finetuning with sufficient multilingual data can compensate for a weaker encoder. At 9B scale, both converge to strong performance. In short, multilingual-aware encoders provide an advantage when data is scarce, but extensive multilingual training can close the gap.

\textbf{High-quality English captions are enough to ensure strong alignment.}
To assess whether multilingual supervision is necessary during alignment pretraining, we train two versions of \towervision{} on both scales, 2B and 9B. 

\begin{wraptable}[11]{r}{8.1cm}
\vspace{-0.7cm}
\centering
\small
    \caption{Effect of using multilingual versus English-only captions during projector pretraining on ALM-Bench. Results indicate low to no gains from adding multilingual data at this stage.}
    \vspace{0.2cm}
    \centering
    \small
    \resizebox{0.585\textwidth}{!}{%
    \begin{tabular}{lcccc}
    \toprule
    \textbf{\towervision{}} & \multicolumn{2}{c}{\textbf{2B}} & \multicolumn{2}{c}{\textbf{9B}}\\
    \cmidrule(lr){2-3} \cmidrule(lr){4-5}
     \textbf{Projector} & \textbf{En} & \textbf{Multi } & \textbf{En} & \textbf{Multi } \\
    \midrule
    En &  77.1 & \textbf{81.1} & \textbf{83.6} & \textbf{85.2} \\
    En+Multi  & \textbf{77.9} & 79.3 & 83.0 & 84.1 \\
    \bottomrule
    \end{tabular}
    }
\label{tab:projector_multilinguality}
\end{wraptable} 
The first version uses only English-only captions from \textsc{PixMo-Cap}, comprising $702,205$ text-image pairs. The second version uses the same English captions combined with a high-quality translated subset from \textsc{PIXMO-CAP}, where data was uniformly translated into the supported languages as described in \S\ref{sec:visionblocks}, comprising 367,779 samples.
We evaluate the models in \textsc{ALM-Bench} to measure \towervision{} performance both in English and across multiple non-English languages, providing insights into how well cross-lingual generalization is preserved or improved.
As shown in Table \ref{tab:projector_multilinguality}, adding high-quality multilingual captions during the projector alignment stage has little to no positive effect and, in some cases, slightly decreases performance on the multilingual subset. This suggests that the most effective strategy is to focus on diverse and high-quality captions, ensuring strong alignment between visual and textual modalities, rather than prioritizing extensive multilingual coverage at this stage.

\begin{figure}[t]
    \centering
    \includegraphics[width=\textwidth]{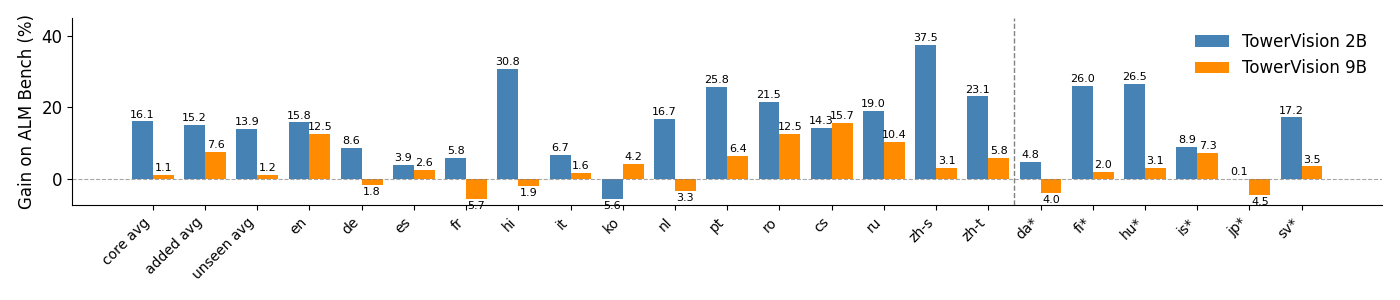}
    \vspace{-0.8cm}
    \caption{Performance of TowerVision models on 10 vs 20 languages/dialects at 2B and 9B scales. The bars indicate the accuracy gains by training on 20 (all) versus 10 (core) languages (more details in \S\ref{app:cross-ling-gen}). Unsupported languages are marked with $^*$.}
    \label{fig:gain_loss}
\end{figure}

\textbf{Expanding languages improves cross-lingual generalization in VLMs.} \label{tab:xcross_gen} 
We study how language coverage in training data impacts performance on both included and excluded languages. Specifically, we compare training on 10 high-resource “core languages” versus the full set of languages, while controlling for dataset size. Our questions are: (i) whether adding balanced multimodal data for more languages improves performance on core languages \citep{conneau2020unsupervisedcrosslingualrepresentationlearning,hu2020xtrememassivelymultilingualmultitask}, and (ii) whether unsupported languages benefit in zero-shot fashion if related languages are present \citep{ni2021m3plearninguniversalrepresentations}. We train \towervision{} at 2B and 9B scales using the recipe in \S\ref{subsec:recipe}, first on 10 ``core'' languages (English, German, Dutch, Portuguese, Russian, Simplified and Traditional Chinese, Spanish, French, Italian), then on all available languages. Results in Figure~\ref{fig:gain_loss} (more details in \S\ref{app:cross-ling-gen}) show that broader language coverage consistently improves performance, with larger gains at the 2B scale. Zero-shot improvements for unsupported languages further support cross-lingual transfer when related languages are included. These findings highlight the value of expanding multilingual data, particularly for smaller models.

\begin{table*}[t]
\centering
\small
\caption{Accuracy (\%) on ViMUL-Bench across 14 languages averaged across multiple-choice and open-ended questions. \underline{Underlined} values mark the best score within \towervision{}/\towervideo{} variants;  Unsupported languages are marked with $^*$.}
\vspace{0.2cm}
\setlength{\tabcolsep}{2pt} 
\resizebox{\textwidth}{!}{%
\begin{tabular}{lcccccccccccccc}
\toprule
\textbf{Model} & \textbf{ar} & \textbf{bn$^*$} & \textbf{zh} & \textbf{en} & \textbf{fr} & \textbf{de} & \textbf{hi} & \textbf{ja} & \textbf{ru} & \textbf{si$^*$} & \textbf{es} & \textbf{sv} & \textbf{ta$^*$} & \textbf{ur$^*$} \\
\midrule
\rowcolor{blue!10} \towervision-2B & 18.9 & \textbf{19.5} & 21.7 & 34.2 & 28.9 & 28.3 & 25.1 & 22.2 & 24.8 & 16.3 & 30.4 & 27.1 & 16.1 & \textbf{19.9} \\
\rowcolor{blue!15} \towervideo-2B (english only) & \textbf{25.7} & 17.8 & 26.7 & \textbf{45.5} & \textbf{42.3} & 34.8 & 27.8 & 27.7 & 34.4 & \textbf{17.9} & \textbf{37.8} & 34.0 & \textbf{18.3} & 19.7  \\
\rowcolor{blue!20} \towervideo-2B (multilingual) & 23.0 & 18.9 & \textbf{35.9} & 45.2 & 39.6 & \textbf{39.7} & \textbf{37.2} & \textbf{34.1} & \textbf{38.0} & 17.1 & 37.4 & \textbf{38.0} & 17.7 & 18.7 \\
\bottomrule
\end{tabular}
}
\label{tab:video_ablation}
\end{table*}

\textbf{How does multilingual data affect video fine-tuning?} 
To assess the impact of our multilingual data (see \S~\ref{sec:visionblocks}) during video fine-tuning, we present results in Table~\ref{tab:video_ablation} for two baselines: (i) the original \towervision{}-2B model and (ii) \towervideo{}-2B trained on the full English-only LLaVA-Video-178k dataset. Fine-tuning with video substantially improves the performance of TowerVision models compared to image-text-only variants, highlighting the importance of temporal information for video-language understanding. Incorporating multilingual data further enhances cross-lingual generalization, while English performance remains largely stable, indicating that adding multiple languages does not compromise primary-language capabilities, even though the multilingual models are trained on substantially less English data.

\section{Conclusion}


We introduced \towervision{}, a suite of multimodal models for image-text and video-text tasks, designed with a strong emphasis on cultural understanding and multilinguality. Our models demonstrate competitive, and in several cases improved, multilingual performance across a range of benchmarks when compared with existing open multimodal systems. Alongside this, we released \visionblocks{}, a high-quality vision-language dataset, and provided a detailed training recipe covering data, encoders, and text backbones, complemented by an extensive ablation study on key components of our approach.

We hope that these contributions---spanning models, data, and methodology---help advance research on culturally diverse multilingual multimodal language models, and accelerate progress toward narrowing the performance gap with English-centric settings.

\section{Acknowledgments}

This work was supported by the project DECOLLAGE (ERC-2022-CoG 101088763), by the Portuguese Recovery and Resilience Plan through project C645008882-00000055 (Center for Responsible AI), and by FCT/MECI through national funds and when applicable co-funded EU funds under UID/50008: Instituto de Telecomunicações. The models were trained at EuroHPC infrastructure MareNostrum5 through grants EHPC-AI-2024A05-044 and 2025.00272.CPCA.A3.

\section{Ethics statement}

This work develops and evaluates multilingual vision-language models using publicly available datasets as well as our own synthetic and translated data. We acknowledge potential risks, including biased model outputs and unintended misuse of generated content. While we have taken steps to ensure diversity and maximum data quality, we always encourage careful evaluation and responsible deployment of these models in real-world scenarios. Our research does not involve sensitive personal data or tasks with direct safety-critical impact.

\section{Reproducibility Statement}

This work provides detailed descriptions of the data, model architectures, training procedure (including the codebase), and evaluation benchmarks used. All datasets used are either publicly available or created by our team (synthetic and translated), with the respective system prompts shared for maximum transparency. Additionally \towervision{} all the collection of models, code for data preprocessing, training, and evaluation will be released to facilitate replication of our results. We aim to ensure that other researchers can reproduce our findings with minimal effort.


\bibliography{iclr2026_conference}
\bibliographystyle{iclr2026_conference}

\appendix
\section{Appendix}

\subsection{Full Description of \visionblocks{}}\label{sec:app_visionblocks}

Table~\ref{tab:vision_blocks_full} shows the full details and statistics of the \visionblocks{} dataset. 

\begin{table}[t]
    \caption{Overview of dataset composition across categories. Each dataset lists its sample size with the proportion of the total in parentheses, along with its collection type tag (\humantag{}, \synthetictag{}, or \translatedtag{}). Totals are shown for English-only and Multilingual subsets, as well as the overall dataset size.}
    \vspace{0.2cm}
    \centering
    \small
    \setlength{\tabcolsep}{2pt} 
    \begin{tabular}{llrc}
    \toprule
    \textbf{Category} & \textbf{Dataset} & \textbf{Samples (\%)} & \textbf{Tag} \\
    \midrule

    Chart/Plot
     & DVQA & 199{,}995 (3.17\%) & \humantag \\
     & ChartQA & 25{,}055 (0.40\%) & \synthetictag \\
     & PlotQA & 157{,}070 (2.49\%) & \humantag \\
     & TabMWP & 22{,}717 (0.36\%) & \humantag \\
    \cmidrule(lr){2-4}

    General VQA
     & VQAv2 & 428{,}708 (6.79\%) & \humantag \\
     & RLAIF-4V & 59{,}408 (0.94\%) & \synthetictag \\
    \cmidrule(lr){2-4}

    Doc VQA
     & DocVQA & 9{,}664 (0.15\%) & \synthetictag \\
     & TextVQA & 15{,}690 (0.25\%) & \synthetictag \\
     & ST-VQA & 17{,}242 (0.27\%) & \humantag \\
     & PixMo-Docs & 3{,}634 (0.06\%) & \humantag \\
    \cmidrule(lr){2-4}

    Reasoning/Knowledge
     & A-OKVQA & 11{,}853 (0.19\%) & \synthetictag \\
     & OKVQA & 9{,}009 (0.14\%) & \humantag \\
     & AI2D & 7{,}791 (0.12\%) & \humantag \\
     & ScienceQA & 758 (0.012\%) & \humantag \\
    \cmidrule(lr){2-4}

    Multilingual/Cultural
     & Pangea-Cultural & 55{,}438 (0.88\%) & \humantag \\
     & Pangea-Multi & 428{,}838 (6.79\%) & \humantag \\
     & PixMo-Cap-Translated & 367{,}779 (5.83\%) & \translatedtag \\
     & CulturalGround-OE & 401{,}149 (6.35\%) & \humantag \\
     & CulturalGround-MCQs & 379{,}834 (6.02\%) & \humantag \\
    \cmidrule(lr){2-4}

    Specialized VQA
     & IconQA & 19{,}543 (0.31\%) & \synthetictag \\
     & InfographicVQA & 2{,}049 (0.03\%) & \synthetictag \\
     & Stratos & 12{,}585 (0.20\%) & \humantag \\
    \cmidrule(lr){2-4}

    Counting/Math
     & TallyQA & 98{,}675 (1.56\%) & \humantag \\
     & PixMo-Count & 8{,}128 (0.13\%) & \humantag \\
    \cmidrule(lr){2-4}

    Vision/Text
     & VBlocks-PixMo-AMA & 154{,}336 (2.44\%) & \humantag \\
     & VBlocks-PixMo-Cap & 702{,}205 (11.12\%) & \humantag \\
     & VBlocks-PixMo-CapQA & 262{,}862 (4.16\%) & \humantag \\
     & EuroBlocks-SFT & 1{,}094{,}265 (17.34\%) & \humantag \\
    \cmidrule(lr){2-4}

    Video/Text
     & LLaVA-Video-178k-subset & 697{,}618 (11.05\%) & \humantag \\
     & LLaVA-Video-178k-translated & 697{,}617 (11.05\%) & \translatedtag \\
     
    \midrule
    \multicolumn{2}{r}{\textbf{Total (English)}} & 3{,}982{,}630 (63.1\%) & \\
    \multicolumn{2}{r}{\textbf{Total (Multilingual)}} & 2{,}330{,}656 (36.9\%) & \\
    \multicolumn{2}{r}{\textbf{Overall Total}} & 6{,}313{,}286 (100\%) & \\
    \bottomrule
    \end{tabular}
    \label{tab:vision_blocks_full}
\end{table}

\subsection{Models Checkpoints}\label{app:model_checkpoints}

Table \ref{tab:vlm_checkpoints} lists all model checkpoints used for comparative baselines. We use checkpoints released HuggingFace when possible.
\begin{table}[t]
    \centering
    \setlength{\tabcolsep}{2pt} 
    \begin{tabularx}{\textwidth}{l c X}
        \toprule
        \textbf{Model} & \textbf{Params} & \textbf{Checkpoint Link} \\
        \midrule
        Qwen2.5-VL-Instruct & 3B & \url{https://huggingface.co/Qwen/Qwen2.5-VL-3B-Instruct} \\
        Qwen2.5-VL-Instruct & 7B & \url{https://huggingface.co/Qwen/Qwen2.5-VL-7B-Instruct} \\
        Gemma2-it & 2B & \url{https://huggingface.co/google/gemma-2-2b-it} \\
        Gemma2-pt & 2B & \url{https://huggingface.co/google/gemma-2-2b} \\
        Gemma2-it & 9B & \url{https://huggingface.co/google/gemma-2-9b-it} \\
        Gemma2-pt & 9B & \url{https://huggingface.co/google/gemma-2-9b} \\
        Gemma3-it & 4B & \url{https://huggingface.co/google/gemma-3-4b-it} \\
        Gemma3-it & 12B & \url{https://huggingface.co/google/gemma-3-12b-it} \\
        CulturalPangea & 7B & \url{https://huggingface.co/neulab/CulturalPangea-7B} \\
        LLaVA-Next & 7B & \url{llava-hf/llava-v1.6-mistral-7b-hf} \\
        Aya-Vision & 8B & \url{https://huggingface.co/CohereForAI/aya-vision-8b} \\
        Pixtral & 12B & \url{https://huggingface.co/mistralai/Pixtral-12B-2409} \\
        Phi-4-Multimodal & 14B & \url{https://huggingface.co/microsoft/Phi-4-multimodal-instruct} \\
        \bottomrule
    \end{tabularx}
    \caption{\textbf{Model checkpoints.} Parameters and HuggingFace links for models included in our evaluation suite.}
    \label{tab:vlm_checkpoints}
\end{table}

\subsection{Vision Encoder Variants}
\label{cap:tiles_patch_vision_encoders}

Beyond selecting a more multilingual vision encoder, several other factors significantly influence its performance. These include the input image resolution supported by the encoder, the number of patches it uses, which determines the total number of visual tokens for a given image resolution (e.g, for an img resolution of $224\times224$ using patch size of 14 we obtain 256 visual tokens) and the number of tiles.

Our goal is to empirically identify the optimal configuration for processing visual inputs, focusing on these three factors.

Specifically, we perform experiments using the \towervision{} 2B version with variants of \textsc{SigLIP2} framework:
\begin{enumerate}
    \item Image resolution: We vary the input image size between $384\times384$, $224\times224$, and $512\times512$ to examine its effect on feature extraction quality. 
    \item Patch numbers: We test different patch sizes (14 and 16) to assess how granularity impacts the learned representations. Smaller patches capture finer details but increase the number of tokens, affecting the context length the model must handle.
    \item Number of tiles: Beyond the default 6 tiles, we also experiment with 4 and 22 tiles. The number of tiles is adjusted to the image resolution: lower-resolution images (e.g, $224\times224$) require more tiles to cover the same amount of visual information as a higher-resolution encoder (e.g., $512\times512$). For example, an image with resolution ($1024,1024$) processed with a $512\times512$ encoder requires roughly 4 tiles to cover the full image, whereas a $224\times224$ encoder would need at least 25 tiles (including padding) to achieve similar coverage.
    This creates a trade-off between capturing detailed local information and maintaining manageable context length.
\end{enumerate}

These experiments allow us to systematically compare variations while keeping other components constant, providing insights into which configuration yields the best overall performance. Results are reported in Table \ref{cap:vision_encoder_app}, highlighting the trade-offs between resolution, patch granularity, and style diversity.

\begin{table}[t]
\centering
\small
\caption{\textbf{Impact of Vision Encoder Configuration and Instruction Tuning.} 
Evaluation of \towerp{} models across English and multilingual tasks with varying image resolution, patch size, and number of tiles. Results highlight how these design choices affect overall performance.}
\vspace{0.2cm}

\resizebox{\textwidth}{!}{
\setlength{\tabcolsep}{8pt} 

\begin{tabular}{lccc|cccc}
\toprule
\textbf{Resolution} & \textbf{Patch Size} & \textbf{Tiles} & \multicolumn{2}{c}{\textbf{English}} & \multicolumn{2}{c}{\textbf{Multilingual}} \\
\cmidrule(lr){4-5} \cmidrule(lr){6-7}
 & & & TextVQA & OCRBench & CC-OCR & ALM-Bench \\
\midrule
224x224 & 14 & 22 & 59.1 & 53.3 & 37.2 & 70.5 \\
224x224 & 16 & 20 & 68.6 & 57.8 & 44.3 & 75.2 \\
384x384 & 14 & 6 & \textbf{70.3} & \textbf{62.1} & \textbf{46.1} & \textbf{75.6} \\
512x512 & 16 & 4 & 64.0 & 55.7 & 39.6 & 74.7 \\
\bottomrule
\end{tabular}
}
\label{cap:vision_encoder_app}
\end{table}

\subsection{Cross-Lingual Generalization}
\label{app:cross-ling-gen}

\begin{table}[t]
\centering
\small

\caption{Cross-lingual performance of \towervision{} models at 2B and 9B scales. \textit{Core Langs} refers to a set of 10 languages: English, German, Dutch, Portuguese, Russian, Simplified and Traditional Chinese, Spanish, French and Italian. \textit{Core+Added Langs} includes all languages supported by \towervision{}. \textit{Unseen} languages are those not encountered during training.
Bold values indicate the better result within each scale. Positive gains from adding languages are highlighted in light green, negative gains in light red. Overall, adding more languages tends to improve performance across the board, demonstrating strong cross-lingual transfer capabilities, even for unseen languages.}
\vspace{0.2cm}

\resizebox{\textwidth}{!}{
\setlength{\tabcolsep}{4pt} 
\begin{tabular}{lccc|ccc}
\toprule
\textbf{} & \multicolumn{3}{c|}{\textbf{TowerVision-2B}} & \multicolumn{3}{c}{\textbf{TowerVision-9B}} \\
\cmidrule(lr){2-4} \cmidrule(lr){5-7}
\textbf{Metric / Lang} & Core Langs & Core + Added Langs & Gain & Core Langs & Core + Added Langs & Gain \\
\midrule
English (en) & 60.9 & \textbf{76.6} & \textcolor{lightgreen}{+15.8} & 70.3 & \textbf{82.8} & \textcolor{lightgreen}{+12.5} \\
Core Avg & 65.3 & \textbf{81.3} & \textcolor{lightgreen}{+16.1} & 81.5 & \textbf{82.6} & \textcolor{lightgreen}{+1.1} \\
Added Avg & 60.2 & \textbf{75.4} & \textcolor{lightgreen}{+15.2} & 76.3 & \textbf{84.3} & \textcolor{lightgreen}{+7.6} \\
Unseen Avg & 69.2 & \textbf{83.0} & \textcolor{lightgreen}{+13.9} & 81.2 & \textbf{82.5} & \textcolor{lightgreen}{+1.2} \\
\midrule
German (de) & 75.9 & \textbf{84.5} & \textcolor{lightgreen}{+8.6} & \textbf{89.7} & 87.9 & \textcolor{lightred}{-1.8} \\
Spanish (es) & 56.6 & 60.5 & \textcolor{lightgreen}{+3.9} & 73.7 & \textbf{76.3} & \textcolor{lightgreen}{+2.6} \\
French (fr) & 76.9 & \textbf{82.7} & \textcolor{lightgreen}{+5.8} & \textbf{86.5} & 80.8 & \textcolor{lightred}{-5.7} \\
Hindi (hi) & 44.2 & \textbf{75.0} & \textcolor{lightgreen}{+30.8} & 82.7 & 80.8 & \textcolor{lightred}{-1.9} \\
Italian (it) & 75.0 & \textbf{81.7} & \textcolor{lightgreen}{+6.7} & \textbf{96.7} & \textbf{98.3} & \textcolor{lightgreen}{+1.6} \\
Korean (ko) & \textbf{76.4} & 70.8 & \textcolor{lightred}{-5.6} & 75.0 & \textbf{79.2} & \textcolor{lightgreen}{+4.2} \\
Dutch (nl) & 70.0 & \textbf{86.7} & \textcolor{lightgreen}{+16.7} & \textbf{90.0} & 86.7 & \textcolor{lightred}{-3.3} \\
Portuguese (pt) & 64.5 & \textbf{90.3} & \textcolor{lightgreen}{+25.8} & 85.5 & \textbf{91.9} & \textcolor{lightgreen}{+6.4} \\
Romanian (ro) & 58.9 & \textbf{80.4} & \textcolor{lightgreen}{+21.5} & 75.0 & \textbf{87.5} & \textcolor{lightgreen}{+12.5} \\
Czech (cs) & 61.4 & \textbf{75.7} & \textcolor{lightgreen}{+14.3} & 74.3 & \textbf{90.0} & \textcolor{lightgreen}{+15.7} \\
Russian (ru) & 65.5 & \textbf{84.5} & \textcolor{lightgreen}{+19.0} & 65.5 & \textbf{75.9} & \textcolor{lightgreen}{+10.4} \\
Chinese (simp.) (zh-hans) & 50.0 & \textbf{87.5} & \textcolor{lightgreen}{+37.5} & 68.8 & 71.9 & \textcolor{lightgreen}{+3.1} \\
Chinese (trad.) (zh-hant) & 53.8 & \textbf{76.9} & \textcolor{lightgreen}{+23.1} & 61.5 & 67.3 & \textcolor{lightgreen}{+5.8} \\
\midrule
Danish (da)* & 66.1 & 70.9 & \textcolor{lightgreen}{+4.8} & \textbf{90.3} & 86.3 & \textcolor{lightred}{-4.0} \\
Finnish (fi)* & 56.0 & \textbf{82.0} & \textcolor{lightgreen}{+26.0} & 70.0 & 72.0 & \textcolor{lightgreen}{+2.0} \\
Hungarian (hu)* & 68.8 & \textbf{95.3} & \textcolor{lightgreen}{+26.5} & 79.7 & \textbf{82.8} & \textcolor{lightgreen}{+3.1} \\
Icelandic (is)* & 67.6 & \textbf{76.5} & \textcolor{lightgreen}{+8.9} & 76.5 & \textbf{83.8} & \textcolor{lightgreen}{+7.3} \\
Japanese (jp)* & \textbf{78.8} & 78.9 & \textcolor{lightgreen}{0.1} & \textbf{84.8} & 80.3 & \textcolor{lightred}{-4.5} \\
Swedish (sv)* & 77.6 & \textbf{94.8} & \textcolor{lightgreen}{+17.2} & 86.2 & \textbf{89.7} & \textcolor{lightgreen}{+3.5} \\
\bottomrule
\end{tabular}
}
\label{tab:crosslingual_generalization}
\end{table}

\subsection{System Prompts}

\subsubsection{Tower System Prompts used for Translation}
\label{cap:tower_translation_captions}

The prompts vary in style and specificity to improve diversity and capture nuanced meaning from the original English captions. They are grouped by language and include multiple phrasings for the same instruction to encourage robust translations.

\begin{small}
\begin{verbatim}
# English prompts
EN_PROMPTS = [
    "Describe this image.",
    "What can you see in this picture?",
    "Tell me what's in this image.",
    "Explain what this image shows.",
    "Caption this image.",
    "What's happening in this picture?",
    "Provide a description of this image."
]

# European Portuguese prompts
PT_PROMPTS = [
    "Descreva esta imagem.",
    "O que consegue ver nesta fotografia?",
    "Diga-me o que está nesta imagem.",
    "Explique o que esta imagem mostra.",
    "Legende esta imagem.",
    "O que se passa nesta fotografia?",
    "Forneça uma descrição desta imagem."
]

# French prompts
FR_PROMPTS = [
    "Décrivez cette image.",
    "Que pouvez-vous voir sur cette photo?",
    "Dites-moi ce qu'il y a dans cette image.",
    "Expliquez ce que cette image montre.",
    "Légendez cette image.",
    "Que se passe-t-il sur cette photo?",
    "Fournissez une description de cette image."
]

# Dutch prompts
NL_PROMPTS = [
    "Beschrijf deze afbeelding.",
    "Wat zie je op deze foto?",
    "Vertel me wat er op deze afbeelding staat.",
    "Leg uit wat deze afbeelding laat zien.",
    "Onderschrift deze afbeelding.",
    "Wat gebeurt er op deze foto?",
    "Geef een beschrijving van deze afbeelding."
]

# German prompts
DE_PROMPTS = [
    "Beschreiben Sie dieses Bild.",
    "Was können Sie auf diesem Foto sehen?",
    "Sagen Sie mir, was auf diesem Bild zu sehen ist.",
    "Erklären Sie, was dieses Bild zeigt.",
    "Beschriften Sie dieses Bild.",
    "Was passiert auf diesem Foto?",
    "Geben Sie eine Beschreibung dieses Bildes."
]

# Spanish prompts
ES_PROMPTS = [
    "Describe esta imagen.",
    "¿Qué puedes ver en esta foto?",
    "Dime qué hay en esta imagen.",
    "Explica qué muestra esta imagen.",
    "Pon un título a esta imagen.",
    "¿Qué está pasando en esta foto?",
    "Proporciona una descripción de esta imagen."
]

# Italian prompts
IT_PROMPTS = [
    "Descrivi questa immagine.",
    "Cosa puoi vedere in questa foto?",
    "Dimmi cosa c'è in questa immagine.",
    "Spiega cosa mostra questa immagine.",
    "Dai un titolo a questa immagine.",
    "Cosa sta succedendo in questa foto?",
    "Fornisci una descrizione di questa immagine."
]


\end{verbatim}
\end{small}

\begin{CJK*}{UTF8}{mj}
\begin{verbatim}
# Korean prompts
KO_PROMPTS = [
    "이 이미지를 설명해주세요.",
    "이 사진에서 무엇을 볼 수 있나요?",
    "이 이미지에 무엇이 있는지 알려주세요.",
    "이 이미지가 보여주는 것을 설명해주세요.",
    "이 이미지에 캡션을 달아주세요.",
    "이 사진에서 무슨 일이 일어나고 있나요?",
    "이 이미지에 대한 설명을 제공해주세요."
]

\end{verbatim}
\end{CJK*}

\begin{CJK*}{UTF8}{gbsn}
\begin{verbatim}
# Chinese prompts
ZH_PROMPTS = [
    "描述这张图片。",
    "你能在这张照片中看到什么？",
    "告诉我这张图片里有什么。",
    "解释这张图片展示了什么。",
    "为这张图片添加说明。",
    "这张照片中发生了什么？",
    "提供这张图片的描述。"
]
\end{verbatim}
\end{CJK*}

\subsubsection{Gemini 2.5 System Prompts}
\label{cap:gemini_synthetic_captions}

We generate synthetic captions using the Gemini 2.5 API with a diverse set of system prompts. These prompts are designed to produce varied response formats, including direct answers, caption-plus-answer pairs, and structured final-answer formats.

\begin{small}
\begin{verbatim}
# Direct answer formats
    "Answer the question concisely.",
    "Provide a brief, direct answer to the question.",
    "Keep your response short and to the point.",
    "Give a concise answer based on what you see in the image.",
    "Answer directly based on the visual information.",
    "Respond with a short, clear answer to the question.",
    "Be brief and direct in your response."

# Simple caption + answer formats
    "First provide a caption of what you see, then give your answer.",
    "Write a brief caption describing the image, followed by your answer to the question.",
    "Start with a description of the image, then provide your answer clearly marked as 'Answer:'.",
    "First write 'Caption: <brief image description>' then answer the question.",
    "Begin with 'Caption: [what you see in the image]' followed by your response to the question.",
    "Start by writing 'CAPTION: {description}' before answering the question."

# Final Answer formats
    "End your response with 'Final Answer: <your answer>'.",
    "Conclude with 'Final Answer: <your answer>'.",
    "After looking at the image, provide 'Final Answer: <your answer>'.",
    "Your response should end with 'Final Answer: <your answer>'.",
    "First describe what you see, then provide 'Final Answer: <your answer>'.",
    "Always end your response with 'Final Answer: <your answer>' after analyzing the image.",
    "Provide a concise answer. End with 'Final Answer: <your answer>'."
    
# Naive formats (simple, direct)
    "Describe the image and answer the question.",
    "Begin by describing the image and then answer the question.",
    "Provide a brief description of the image and then answer the question.",
    "Answer the question in a helpful and informative manner.",
    "Start by describing the image and then answer the question.",
    "You are a helpful assistant. Describe the image and answer the question."

# Simple formatted caption/answer pairs
    "Caption: <description> → Answer: <response>",
    "Image shows: <description> | My answer: <response>",
    "[CAPTION] <description> [ANSWER] <response>",
    "# Image: <description>\n# Answer: <response>",
    "First 'Image Description: <what you see>' then 'Answer: <your response>'"

# With specific markers
    "<description><answer>",
    "Image: <description> → Answer: <conclusion>",
    "<IMAGE> describe what you see </IMAGE> <ANSWER> provide your response </ANSWER>"
"Begin with '{IMAGE DESCRIPTION}' and end with '{FINAL ANSWER}'."
\end{verbatim}
\end{small}

These prompts are used to generate high-quality captions that improve instruction-following and visual description diversity.


\end{document}